\documentclass[10pt,twocolumn,letterpaper]{article}

\usepackage{pdfpages}
\usepackage{cvpr}
\usepackage{times}
\usepackage{epsfig}
\usepackage{graphicx}
\usepackage{amsmath}
\usepackage{amssymb}
\usepackage{hhline}
\usepackage[table,x11names,dvipsnames,table]{xcolor}
\usepackage{bbm}
\usepackage{enumitem}
\usepackage{algorithm}
\usepackage[noend]{algpseudocode}
\usepackage{url}

\makeatletter
\def\BState{\State\hskip-\ALG@thistlm}
\makeatother

\newcommand{\Exp}{{\rm I\kern-.3em E}}
\definecolor{myorchid}{RGB}{150,10,30}
\definecolor{myblue}{RGB}{10,30,250}
\definecolor{mygreen}{RGB}{10,120,10}
\definecolor{myyellow}{RGB}{220,220,10}

\newcommand{\karren}[1]{\color{myorchid}{KY: #1}\color{black}}

\renewcommand{\paragraph}[1]{\vspace{0.03cm} \noindent{\bf #1}}
\usepackage{multirow}
\renewcommand{\vec}[1]{\mathbf{#1}}

\usepackage[pagebackref=true,breaklinks=true,letterpaper=true,colorlinks,bookmarks=false]{hyperref}

\cvprfinalcopy 

\def\cvprPaperID{7333} 

\ifcvprfinal\fi
\begin{document}


\title{Defending Multimodal Fusion Models against Single-Source Adversaries}

\author{Karren Yang\textsuperscript{1}
\quad
Wan-Yi Lin\textsuperscript{2}
\quad
Manash Barman\textsuperscript{3}
\quad
Filipe Condessa\textsuperscript{2}
\quad
Zico Kolter\textsuperscript{2,3}\\

\textsuperscript{1}Massachusetts Institute of Technology
\quad
\textsuperscript{2}Bosch Center for AI\thanks{KY and MB completed work during an internship at BCAI. Work is sponsored by DARPA (Grant number HR11002020006).}
\quad
\textsuperscript{3}Carnegie Mellon University\\
{\tt\small karren@mit.edu}
,
{\tt\small wan-yi.lin@us.bosch.com}
,
{\tt\small mbarman@andrew.cmu.edu}
\\
{\tt\small filipe.condessa@us.bosch.com}
,
{\tt\small zkolter@cs.cmu.edu}
}

\maketitle
\thispagestyle{empty}

\begin{abstract}
Beyond achieving high performance across many vision tasks, multimodal models are expected to be robust to single-source faults due to the availability of redundant information between modalities. In this paper, we investigate the robustness of multimodal neural networks against worst-case (\ie, adversarial) perturbations on a single modality. We first show that standard multimodal fusion models are vulnerable to single-source adversaries: an attack on any single modality can overcome the correct information from multiple unperturbed modalities and cause the model to fail. This surprising vulnerability holds across diverse multimodal tasks and necessitates a solution.
Motivated by this finding, we propose an adversarially robust fusion strategy that trains the model to compare information coming from all the input sources, detect inconsistencies in the perturbed modality compared to the other modalities, and only allow information from the unperturbed modalities to pass through. Our approach significantly improves on state-of-the-art methods in single-source robustness, achieving gains of 7.8-25.2\% on action recognition, 19.7-48.2\% on object detection, and 1.6-6.7\% on sentiment analysis, without degrading performance on unperturbed (\ie, clean) data.
\end{abstract}


\section{Introduction}
\label{sec:introduction}

\begin{figure}
    \centering
    \hspace*{-.5cm}
    \includegraphics[scale=0.5]{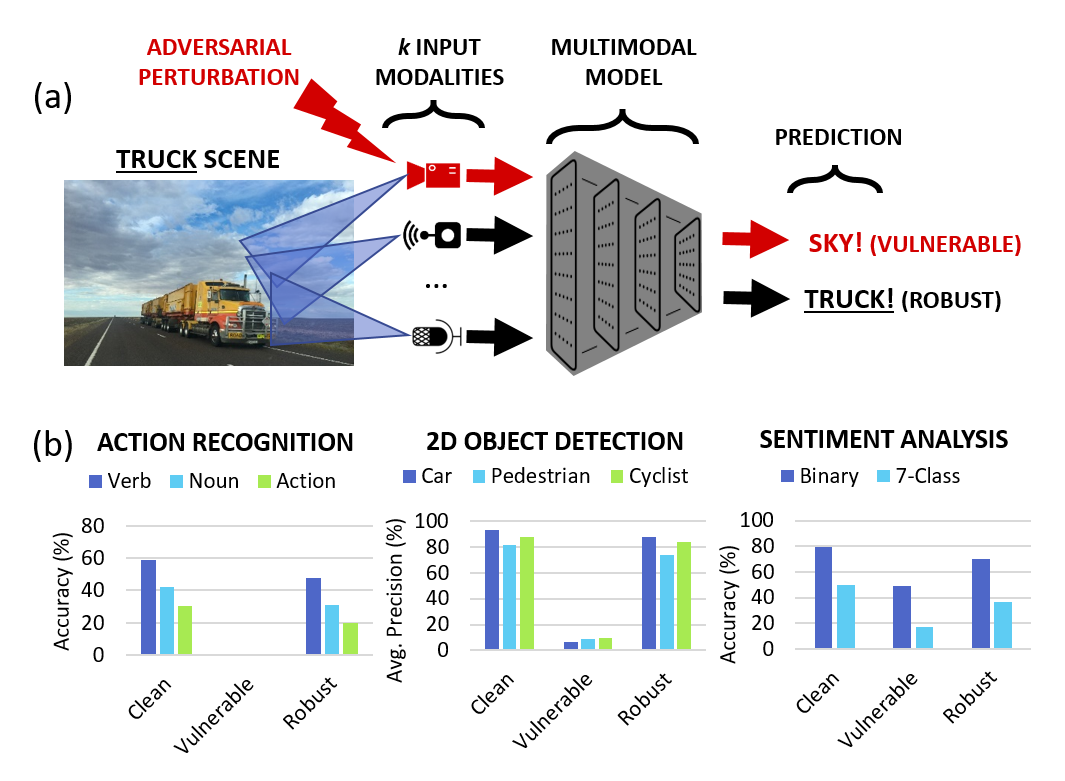}
    \caption{ (a) Example of a single source, worst-case (\ie, adversarial) perturbation on a multimodal model. 
    (b) Standard multimodal models are vulnerable to worst-case perturbations on any single modality (``Vulnerable''). Our  adversarially robust fusion strategy (``Robust'') leverages multimodal consistency to defend against such perturbations without degrading clean performance.}
    \label{fig:single-source-adv-perturb}
\end{figure}


Consider a multimodal neural network, illustrated in Figure \ref{fig:single-source-adv-perturb}(a), that fuses inputs from $k$ different sources to identify objects for an autonomous driving system. If one of the modalities (\eg, RGB) receives a worst-case or adversarial perturbation, does the model fail to detect the truck in the scene? Or does the model make a robust prediction using the remaining $k-1$ unperturbed modalities (\eg, LIDAR, audio, \etc.)? This example illustrates the importance of \emph{single-source adversarial robustness} \cite{kim2019single} for avoiding catastrophic failures in real-world multimodal systems. In a realistic setting, any single modality may be affected by a worst-case perturbation, whereas multiple modalities usually do not fail simultaneously
particularly if the physical sensors are not coupled.
Since multimodal models are being increasingly developed for real-world vision tasks \cite{chen2017multi, wagner2016multispectral, qi2018frustum, ku2018joint}, it is imperative to investigate whether they are robust to worst-case errors that may affect any single modality and, if they are not, to develop strategies to improve robustness.

Despite the importance of this problem, we found to the best of our knowledge that empirical studies of single-source adversarial robustness are lacking. Previous empirical works on multimodal robustness have so far only considered single-source corruptions (\eg, dropout, blurring, \etc.) \cite{kim2018robust, kim2018robusta, kim2019single}, and although Kim \& Ghosh \cite{kim2019single} formulate the problem for the adversarial setting, they do not perform an empirical study. 
In the field of adversarial robustness, most studies have focused on the unimodal setting rather than the multimodal setting \cite{kurakin2016,madry2017towards}. An effective strategy for defending unimodal models against adversaries is \emph{adversarial training} (\ie, end-to-end training of the model on adversarial examples). In principle, adversarial training could be extended to multimodal models as well, but it has several downsides: (1) it is resource-intensive \cite{shafahi2019adversarial} and may not scale well to large, multimodal models that contain many more parameters than their unimodal counterparts; (2) it significantly degrades performance on clean data \cite{madry2017towards}. For these reasons, end-to-end adversarial training may not be practical for multimodal systems used in real-world tasks. 

\paragraph{Contributions.} This paper presents, to our knowledge, the first empirical study of single-source adversarial robustness in multimodal systems. Our contributions are two-fold. 

\noindent (1) We investigate multimodal robustness against single-source adversaries on diverse benchmark tasks with three modalities ($k=3$): action recognition on EPIC-Kitchens \cite{Damen2018EPICKITCHENS}, object detection on KITTI \cite{KITTI_OD}, and sentiment analysis on CMU-MOSI \cite{zadeh2016mosi}. We find that standard multimodal fusion practices are vulnerable to single-source adversarial perturbations, even when there are multiple unperturbed modalities that could yield a correct prediction; naive ensembling of features from a perturbed modality with features from clean modalities does not automatically yield robust prediction. As shown in Figure \ref{fig:single-source-adv-perturb}(b), a worst-case input at any single modality of a multimodal model can outweigh the other modalities and cause the model to fail. In fact, contrary to expectations, a multimodal model ($k=3$) under a single-source perturbation does not necessarily outperform a unimodal model ($k=1$) under the same attack. 

\noindent (2) We propose an adversarially robust fusion strategy that can be applied to mid- to late- fusion models to defend against this vulnerability without degrading clean performance. 
Inspired by recent works 
that detect correspondence between inputs to defend against image manipulation \cite{huh2018fighting}, we hypothesize that a multimodal model can be trained to detect correspondence (or lack thereof) between features from different modalities and use this information to perform a robust feature fusion that defends against the perturbed modality. Our approach extends existing work on adaptive gating strategies \cite{kim2018robust, kim2018robusta, valada2017adapnet, mees2016choosing} with a robust fusion training procedure based on odd-one-out learning \cite{fernando2017self} to improve single-source adversarial robustness without degrading clean performance. Through extensive experiments, we demonstrate that our approach is effective even against adaptive, white-box attacks with access to the robust fusion strategy. 
We significantly outperform state-of-the-art methods in single-source robustness \cite{kim2018robust, kim2018robusta, kim2019single}, achieving gains of 7.8-25.2\% on action recognition on EPIC-Kitchens, 19.7-48.2\% on 2D object detection on KITTI, and 1.6-6.7\% sentiment analysis on CMU-MOSI.

Overall, this paper demonstrates that multimodal models are not inherently robust to single-source adversaries, but that we can improve their robustness without the downsides associated with end-to-end adversarial training in unimodal models. The combination of robust fusion architectures with robust fusion training may be a practical strategy for defending real-world systems against adversarial attacks and establishes a promising direction for future research.

\begin{figure*}[t]
    \centering
    \hspace*{-.8cm}
    \includegraphics[scale=0.21]{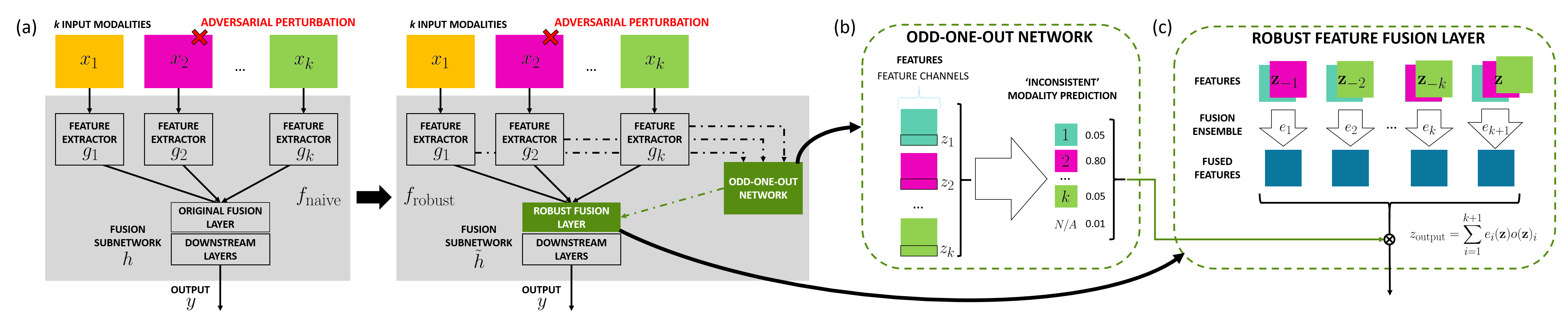}
    \caption{We propose a robust multimodal fusion strategy based on ``odd-one-out'' learning, an auxiliary self-supervised task in which a model is presented with multiple elements and must predict which one of them is different from the others. A multimodal model augmented with an odd-one-out network can be trained to compare information coming from all the input sources, detect the perturbed modality because it is inconsistent with the other modalities, and only allow information from the unperturbed modalities to pass through.}
    \label{fig:moe_layer}
\end{figure*}

\subsection{Related work}
\label{sec:related_work}

\paragraph{Adversarial Robustness.} Vision systems based on deep learning models are susceptible to adversarial attacks-- additive, worst-case, and imperceptible perturbations on the inputs that cause erroneous predictions \cite{domingos2004,biggio2013,Szegedy,Goodfellow2014}. A large number of defense methods against adversarial attacks have been proposed, with the two most effective defenses being end-to-end adversarial training \cite{Goodfellow2014,kurakin2016,madry2017towards}, which synthesizes adversarial examples and includes them in training data, and provably robust training, which provides theoretical bounds \cite{wong2017,raghunathan2018} on the performance. However, these methods have primarily focused on the unimodal setting, in which the input is a single image. In contrast to those works, we consider single-source adversarial perturbations in the multimodal setting and leverage consistent information between modalities to improve the robustness of the model's fusion step. Our training procedure is related to adversarial training in the sense that we also use perturbed inputs, but instead of end-to-end training of model parameters, we focus on designing and training the feature fusion in a robust manner. 
This strategy brings benefits from adversarial training, while retaining performance on clean data and significantly reducing the number of parameters that need to be trained on perturbed data.

\paragraph{Multimodal Fusion Models.} Multimodal neural networks have demonstrated remarkable performance across a variety of vision tasks, such as scene understanding \cite{kazakos2019epic}, object detection \cite{wagner2016multispectral, hassan2020learning}, sentiment analysis \cite{zadeh2016mosi, bagher-zadeh-etal-2018-multimodal, zadeh2018memory, kumar2020gated}, speech recognition \cite{afouras2018deep}, and medical imaging \cite{guo2019deep}. In terms of fusion methods, several approaches that use gating networks have been proposed to weigh sources adaptively depending on the inputs \cite{mees2016choosing, valada2017adapnet, miech2017learnable, arevalo2017gated}. These works focus on leveraging multiple modalities to improve clean performance on the task and do not evaluate or extend these approaches to improve single-source robustness, which is our focus. 

\paragraph{Single Source Robustness.} Several recent works provide important insights into the effects of single-source corruptions such as occlusions, dropout, and Gaussian noise on object detection systems with two modalities ($k=2$) \cite{kim2018robust,kim2018robusta,kim2019single}. In contrast to their work, we consider single-source adversarial perturbations, which explore worst-case failures of multimodal systems due to one perturbed modality. We consider other tasks in addition to object detection and evaluate models with three modalities ($k=3$), in which there are more clean sources than perturbed sources. In terms of defense strategies, robust multimodal fusion methods based on end-to-end robust training \cite{kim2019single} and adaptive gating fusion layers \cite{kim2018robust,kim2018robusta} have been developed to improve single-source robustness to corruptions. We extend this line of work by developing a robust fusion strategy that leverages correspondence between unperturbed modalities to defend against the perturbed modality, and is effective against more challenging adversarial perturbations. 

\section{Single Source Adversarial Perturbations} \label{sec:single-source-adv}
Let $f: \vec{x} \mapsto y$ denote a multimodal model with $k$ input modalities (\ie, $\vec{x} = [x_1, \cdots, x_k]$). We aim to understand the extent to which the performance of $f$ is degraded by worst-case perturbations on any single modality $i \in [k]$ (where $[k] = \{ 1, \cdots, k\}$) while the other $k-1$ modalities 
remain unperturbed. 
To this end, we define a \emph{single-source adversarial perturbation} against $f$ on modality 
$i$ as,
\begin{multline}\label{eq:adv_perturb} 
    \delta^{(i)}(\vec{x}, y; f) := \arg\max_{||\delta||_p \leq \epsilon} \mathcal{L}(f(x_i + \delta, \vec{x_{-i}}), y),
\end{multline}
where $\mathcal{L}$ is the loss function and $\epsilon > 0$ defines the allowable range of the perturbation $\delta^{(i)}$. If we assume that the multimodal inputs $\vec{x}$ and outputs $y$ are sampled from a distribution $\mathcal{D}$, then the \emph{single-source adversarial performance} of $f$ with respect to modality $i \in [k]$ is given by,
\begin{equation}\label{eq:robust_performance}
\Exp_{(\vec{x}, y) \sim \mathcal{D}} \max_{||\delta||_p < \epsilon }\left[\mathcal{L}(f(x_i + \delta, \vec{x_{-i}}), y)\right].
\end{equation}
The difference between the performance of $f$ on unperturbed data, \ie, $\Exp_{(\vec{x}, y) \sim \mathcal{D}} \left[\mathcal{L}(f(\vec{x}), y) \right]$, and its single-source adversarial performance specified in \eqref{eq:robust_performance} indicates, on average, the sensitivity of $f$ to its worst-case inputs on modality $i$. Ideally, a multimodal model that has access to multiple input modalities with redundant information should not be sensitive to perturbations on a single input; it should be able to make a correct prediction by leveraging the remaining $k-1$ unperturbed modalities. However, we find across diverse multimodal benchmark tasks that standard multimodal fusion models are surprisingly vulnerable to these perturbations, even though the clean modalities outnumber the perturbed modality. We discuss the experiments and results in later sections; for now, we emphasize that this vulnerability necessitates a solution.

\section{Adversarially Robust Fusion Strategy}\label{sec:method}

Let $f_{\text{naive}}$ be a standard multimodal neural network, pretrained to achieve acceptable performance on unperturbed data, \ie, it minimizes 
$\Exp_{(\vec{x}, y) \sim \mathcal{D}} \left[\mathcal{L}(f_{\text{naive}}(\vec{x}), y) \right]$.
Our robust fusion strategy aims to improve the single-source robustness of $f_{\text{naive}}$ by leveraging the correspondence between the unperturbed modalities to detect and defend against the perturbed modality.
We assume that $f_{\text{naive}}$ has a mid- to late- fusion architecture, consisting of the composition of modality-specific feature extractors $g_1, \cdots, g_k$ applied to their respective modalities 
and a fusion subnetwork $h$:
\begin{equation}\label{eq:multimodal_model}
    f_{\text{naive}}(\vec{x}) := h(g_1({x_1}), g_2({x_2}), \cdots, g_k({x_k})),
\end{equation}
To make $f_{\text{naive}}$ robust, we equip it with an auxiliary odd-one-out network and a robust feature fusion layer in place of the default feature fusion operation, as shown in Figure \ref{fig:moe_layer}(a). Then we perform robust training based on odd-one-out learning \cite{fernando2017self} and adversarial training \cite{madry2017towards} that focuses on these new modules. The odd-one-out network $o$ is trained to detect the inconsistent or perturbed modality when presented with feature representations of different modalities (Section \ref{sec:odd-one-out-learning}). The robust feature fusion layer ensembles different multimodal fusion operations using the output of the odd-one-out network, ensuring that only the modalities that are consistent with each other are passed to the downstream layers (Section \ref{sec:fusion_layer}). We denote the fusion subnetwork $h$ equipped with the robust feature fusion layer as $\tilde{h}$, and we denote the full, augmented multimodal model as $f_{\text{robust}}$, \ie,
    \begin{equation*}\label{eq:robust_multimodal_model}
    f_{\text{robust}}(\vec{x}) := \tilde{h}(g_1({x_1}), g_2({x_2}), \cdots, g_k({x_k}); o(\{g_i(x_i)\}_{i \in [k]}).
    \end{equation*}
Finally, we jointly train the odd-one-out network $o$ and the fusion subnetwork $\tilde{h}$, while keeping the weights and architectures of the feature extractors $g_1, \cdots, g_k$ fixed from $f_{\text{naive}}$ (Section \ref{sec:training_strategy}). 

\subsection{Odd-one-out learning}\label{sec:odd-one-out-learning}
Odd-one-out learning is a self-supervised task that aims to identify the inconsistent element from a set of otherwise consistent elements \cite{fernando2017self}.
To leverage the shared information between modalities, we propose to augment the multimodal model with an odd-one-out network. Given the set of features $\vec{z} = [z_1, \cdots, z_k]$ extracted from the $k$-modality input, the odd-one-out network predicts whether the multimodal features are consistent with each other (\ie, the inputs are all clean), or whether one modality is inconsistent with the others (\ie, some input has been perturbed). To perform this task, the odd-one-out network must compare the features from different modalities, recognize the shared information between them, and detect any modality that is not consistent with the others. For convenience, we take the features to be the final outputs of the feature extractor networks $g_1, \cdots, g_k$ applied to their respective modalities. In principle, though, these features could also come from any of the intermediate layers of the feature extractors. 

Concretely, the odd-one-out network is a neural network $o$ that maps the features $\vec{z}$ to a vector of size $k+1$, as shown in Figure \ref{fig:moe_layer}(b). The $i$-th entry of this vector indicates the probability that modality $i$ has been perturbed, \ie, ${z_i}$ is inconsistent with the other features. The $k+1$-th entry of the vector indicates the probability that none of the modalities are perturbed. The odd-one-out network $o$ is trained to perform odd-one-out prediction by minimizing the following cross-entropy loss:
\begin{align}\label{eq:odd-one-out-loss}
    -\Exp_{\substack{(\vec{x},y) \sim \mathcal{D}\\
    {z_i} = g_i({x_i})}}
    \big[ \log o(\vec{z})_{k+1} + \sum_{i=1}^k \log o({z_i^\ast}, \vec{z_{-i}})_i] ,
\end{align}
where ${z_i^\ast} = g_i\left(x_i^\ast \right)$ is the feature extracted from perturbed input $x_i^\ast$ that we generate during training.


\begin{table*}[]
\centering
\scriptsize
\begin{tabular}{|l|l|l|l|l|l|}
\hline
\multicolumn{1}{|c|}{\textbf{Dataset}} & \textbf{Tasks} & \textbf{Input Modalities} & \textbf{Model} & \textbf{Adversarial Perturbation} & \textbf{Evaluation Metrics} \\ \hline
\textbf{EPIC-Kitchens} \cite{Damen2018EPICKITCHENS} &
  \begin{tabular}[c]{@{}l@{}}Action \\ recognition\end{tabular} &
  \begin{tabular}[c]{@{}l@{}}Visual frames; \\ Motion frames (flow); \\ Audio (spectrogram)\end{tabular} &
  \begin{tabular}[c]{@{}l@{}}Feature extractors: BNInception \cite{kazakos2019epic} (all);\\ Fusion: feed-forward network + temporal pooling;\\ Odd-one-out network: feed-forward network\end{tabular} &
  \begin{tabular}[c]{@{}l@{}}PGD (10-step):\\ $\epsilon$ = 8/256 (vision) \\ $\epsilon$ = 8/256 (motion)\\ $\epsilon$ = 0.8 (audio)\end{tabular} &
  \begin{tabular}[c]{@{}l@{}}Top-1, top-5 accuracy:\\ Verbs, nouns, actions\end{tabular} \\ \hline
\textbf{KITTI} \cite{KITTI_OD}&
  \begin{tabular}[c]{@{}l@{}}2D object\\ detection\end{tabular} &
  \begin{tabular}[c]{@{}l@{}}Visual frame;\\ Depth map (Velodyne);\\ Depth map (stereo image)\end{tabular} &
  \begin{tabular}[c]{@{}l@{}}Feature extractors: Darknet19 \cite{darknet} (all);\\ Fusion: 1 $\times$ 1 conv layer + YOLO \cite{redmon2018yolov3};\\ Odd-one-out network: 1 $\times$ 1 conv net;\end{tabular} &
  \begin{tabular}[c]{@{}l@{}}PGD (10-step):\\ $\epsilon$ = 16/256 (all)\end{tabular} &
  \begin{tabular}[c]{@{}l@{}}Average precision:\\ Cars (\textgreater{}0.7 IoU),\\ Pedestrians (\textgreater{}0.5 IoU), \\ Cyclists (\textgreater{}0.5 IoU)\end{tabular} \\ \hline
\textbf{MOSI} \cite{zadeh2016mosi} &
  \begin{tabular}[c]{@{}l@{}}Sentiment\\ analysis\end{tabular} &
  \begin{tabular}[c]{@{}l@{}}Visual frame;\\ Audio (mel ceptron);\\ Text\end{tabular} &
  \begin{tabular}[c]{@{}l@{}}Feature extractors: FaceNet\cite{schroff2015facenet} +LSTM (vision), \\ MFCC+LSTM (audio), transformer \cite{} (text);\\ Fusion: feed-forward network\\ Odd-one-out network: feed-forward network\end{tabular} &
  \begin{tabular}[c]{@{}l@{}}PGD (10-step):\\ $\epsilon$ = 8/256 (vision)\\ $\epsilon$ = 0.8 (audio)\\ word replacement \cite{text_attack2}, \\ 1-word per sentence (text)\end{tabular} &
  \begin{tabular}[c]{@{}l@{}}Binary accuracy\\ 7-class accuracy\end{tabular} \\ \hline
\end{tabular}
\caption{A summary table of our experimental setups.}
\label{table:experiments}
\end{table*}

\begin{algorithm}[t]

\caption{Robust Training Strategy.}\label{alg:training}
\begin{algorithmic}[1]
\Procedure{GradientUpdate}{}
\State $\ell_{\text{odd}} \gets 0$
\State $\ell_{\text{task}} \gets 0$
\State Sample $\vec{x} = [x_1, \cdots, x_k], y$ from $\mathcal{D}$
\State $\vec{z} = [z_1, \cdots, z_k] \gets [g_1(x_1), \cdots, g_k(x_k)]$
\State $\ell_{\text{odd}} \gets \ell_{\text{odd}} - \log o(\vec{z})_{k+1}$
\State $\ell_{\text{task}} \gets \ell_{\text{task}} + \mathcal{L}\big(h(\vec{z}, o(\vec{z})), y\big)$
\For {$i \in [k]$}
\State $\delta^{(i)} \gets \delta^{(i)}(\vec{x}, y; f_{\text{robust}})$ \quad (Eqn. \ref{eq:adv_perturb})
\State $z_i^\ast \gets g_i(x_i + \delta^{(i)})$
\State $\ell_{\text{odd}} \gets \ell_{\text{odd}} - \log o(z_i^\ast, \vec{z_{-i}})_i$
\State $\ell_{\text{task}} \gets \ell_{\text{task}} + \mathcal{L}\big(h(z_i^\ast, \vec{z_{-i}}, o(z_i^\ast, \vec{z_{-i}})), y\big)$
\EndFor
\State $\ell \gets \ell_{\text{odd}} + \ell_{\text{task}}$
\State Update $o,h$ based on $\nabla \ell$
\EndProcedure
\end{algorithmic}
\end{algorithm}

\subsection{Robust Feature Fusion Layer} \label{sec:fusion_layer}
To integrate the output of the odd-one-out network $o$ into the multimodal model, we propose a feature fusion layer inspired by the mixture-of-experts layer \cite{shazeer2017outrageously}. This layer consists of an ensemble of $k+1$ feature fusion operations $e_1, \cdots, e_{k+1}$, each of which is specialized to exclude one modality, as illustrated in Figure \ref{fig:moe_layer}(c). Formally, each fusion operation takes the multimodal features $\vec{z}$ as input and performs a fusion of a subset of the features as follows:
\begin{align*}
e_i(\vec{z}) = \text{NN} \big( \oplus \textbf{z}_{-i}  \big) ~\forall i \in [k], \quad
e_{k+1}(\vec{z}) = \text{NN} \big( \oplus \textbf{z}  \big),
\end{align*}
where $\oplus$ denotes the concatenation operation and NN stands for a shallow neural network. By definition, $e_i$ is responsible for performing a fusion of features from all the modalities \emph{except} for $i$, and only $e_{k+1}$ fuses features from all the modalities. If feature ${z_i}$ is not consistent with features from the other $k-1$ modalities because it results from a perturbed input, then $e_i$ receives more weight than the other fusion operations based on the output of the odd-one-out network:
\begin{equation}\label{eq:moe}
z_{\text{output}} = \sum_{i=1}^{k+1} e_i(\vec{z}) o(\vec{z})_i,
\end{equation}
We form a robust fusion subnetwork $\tilde{h}$ by equipping the fusion subnetwork $h$ with this robust feature fusion layer. 
Then $\tilde{h}$ and $o$ are trained to optimize clean performance,
\begin{align}\label{eq:clean_loss}
    \Exp_{\substack{(\vec{x},y) \sim \mathcal{D}\\
    {z_i} = g_i({x_i})}}
    \big[\mathcal{L}\big(\tilde{h}(\vec{z}; o(\vec{z})), y\big)\big],
\end{align}
as well as the single-source robust performance,
\begin{align}\label{eq:adv_loss}
    \Exp_{\substack{(\vec{x},y) \sim \mathcal{D}\\
    {z_i} = g_i({x_i})}}
    \big[\mathcal{L}\big(\tilde{h}({z_i^\ast}, \vec{z_{-i}}; o({z_i^\ast}, \vec{z_{-i}})), y\big)\big],
\end{align}
with respect to each modality, where ${z_i^\ast} = g_i\left(x_i^\ast \right)$ is the feature extracted from perturbed input $x_i^\ast$ that we generate during training. Note that one of the arguments into the fusion network $\tilde{h}$ is now the output of $o$. 

\paragraph{Spatiotemporal Dimensions.} Our formulations assume that $z_1, \cdots, z_k$ are one-dimensional feature representations, in which case the odd-one-out network $o$ and fusion operations $e_1, \cdots, e_{k+1}$ can be implemented as shallow fully-connected networks (\eg, two fully-connected layers). In many multimodal models, the features also have spatiotemporal dimensions that are aligned between different modalities, \ie, ${z_i} \in \mathbb{R}^{c_i \times N_1 \times \cdots \times N_d}$, where $c_i$ is the number of feature channels and $N_1 \times \cdots \times N_d$ are the shared spatiotemporal dimensions (\eg, audio and visual features extracted from a video are aligned along the temporal axis, features extracted from different visual modalities are aligned along the spatial axes). In those cases, our odd-one-out network and fusion operations are more efficiently implemented as convolutional neural networks with $1 \times \cdots \times 1$ filters. This enables us to compute the losses in Equations (\ref{eq:odd-one-out-loss}) and (\ref{eq:moe}) in parallel over the spatiotemporal dimensions.



\subsection{Robust Training Procedure} \label{sec:training_strategy} 
The multimodal model $f_{\text{robust}}$, which is equipped with an odd-one-out network $o$ and fusion subnetwork $\tilde{h}$, contains a mechanism to compare information coming from all the input sources, detect that the perturbed modality is inconsistent with the other unperturbed modalities, and only allow information from the unperturbed modalities to pass through. 
During training, we generate perturbed inputs $x_i^\ast$ using the single-source adversarial perturbations from Equation \ref{eq:adv_perturb}, \ie, we let $$x_i^\ast = x_i + \delta^{(i)}(\mathbf{x}, y, f_{\text{robust}}).$$
Note that this adversarial perturbation is generated against $f_{\text{robust}}$. In other words, our approach performs adversarial training of the fusion network and also leverages the adversarial examples to provide self-supervised labels for odd-one-out learning. We optimize the parameters of the odd-one-out network $o$ and the fusion subnetwork $\tilde{h}$ with respect to the losses in Equations (\ref{eq:odd-one-out-loss}), (\ref{eq:clean_loss}), and (\ref{eq:adv_loss}), as shown in Algorithm \ref{alg:training}. We do not retrain the feature extractors $g_1, \cdots, g_k$, which are already pretrained on clean data.

\begin{table*}[t]
\centering
\scriptsize
\begin{tabular}{|l|c|c|c|c|c|c|c|c|c|c|c|c|}
\hline
\textbf{Fusion} &
  \multicolumn{3}{c|}{\textbf{Clean}} &
  \multicolumn{3}{c|}{\begin{tabular}[c]{@{}c@{}}\textbf{Visual} \\ \textbf{Perturbation}\end{tabular}} &
  \multicolumn{3}{c|}{\begin{tabular}[c]{@{}c@{}}\textbf{Motion} \\ \textbf{Perturbation}\end{tabular}} &
  \multicolumn{3}{c|}{\begin{tabular}[c]{@{}c@{}}\textbf{Audio} \\ \textbf{Perturbation}\end{tabular}} \\ \hline
\textbf{} &
  \multicolumn{1}{c|}{\textbf{Verb}} &
  \multicolumn{1}{c|}{\textbf{Noun}} &
  \multicolumn{1}{c|}{\textbf{Action}} &
  \multicolumn{1}{c|}{\textbf{Verb}} &
  \multicolumn{1}{c|}{\textbf{Noun}} &
  \multicolumn{1}{c|}{\textbf{Action}} &
  \multicolumn{1}{c|}{\textbf{Verb}} &
  \multicolumn{1}{c|}{\textbf{Noun}} &
  \multicolumn{1}{c|}{\textbf{Action}} &
  \multicolumn{1}{c|}{\textbf{Verb}} &
  \multicolumn{1}{c|}{\textbf{Noun}} &
  \multicolumn{1}{c|}{\textbf{Action}} \\ \hline
 \textbf{Oracle (Upper Bound)} &
  - &
  - &
  - &
  55.8 &
  31.4 &
  21.9 &
  50.0 &
  37.2 &
  23.8 &  
  53.9 &
  39.2 &
  25.6
 \\ \hhline{|=|=|=|=|=|=|=|=|=|=|=|=|=|}
\textbf{Concat Fusion} &
  59.0 &
  42.1 &
  30.2 &
  0.1 &
  0.0 &
  0.0 &
  0.2 &
  0.0 &
  0.0 &
  0.1 &
  0.0 &
  0.0 \\ \hline
\textbf{Mean Fusion} &
  56.8 &
  40.4 &
  27.6 &
  0.3 &
  0.8 &
  0.0 &
  0.3 &
  0.3 &
  0.0 &
  0.4 &
  0.3 &
  0.0 \\ \hhline{|=|=|=|=|=|=|=|=|=|=|=|=|=|}
\textbf{LEL+Robust \cite{kim2019single}} &
  61.2 &
  \textbf{43.1} &
  30.5 &
  22.3 &
  11.6  &
  6.6 &
  25.4 &
  24.6 &
  12.0 &
  20.4  &
  17.7 &
  8.0 \\ \hline
\textbf{Gating+Robust \cite{kim2018robust, kim2018robusta}} &
  60.9 &
  43.0 &
  30.6 &
  26.0 &
  10.9 &
  6.2 &
  35.9 &
  26.9 &
  14.3 &
  21.3 &
  16.2 &
  7.0 \\ \hline
\textbf{Ours} &
  \textbf{61.5 } &
  42.5 &
  \textbf{31.4} &
  \textbf{48.0} &
  \textbf{24.2} &
  \textbf{16.8} &
  \textbf{48.5} &
  \textbf{35.6} &
  \textbf{22.1} &
  \textbf{46.5} &
  \textbf{33.3} &
  \textbf{22.1} \\ \hhline{|=|=|=|=|=|=|=|=|=|=|=|=|=|}
  \textbf{$\Delta$-Clean} &
  \textbf{2.5} &
  \textbf{0.3} &
  \textbf{1.2} &
  \textbf{47.7} &
  \textbf{23.4} &
  \textbf{16.8} &
  \textbf{48.2} &
  \textbf{35.3} &
  \textbf{22.1} &
  \textbf{46.1} &
  \textbf{33.0} &
  \textbf{22.1} \\ \hline
  \textbf{$\Delta$-Robust} &
  \textbf{0.3 } &
  -0.6 &
  \textbf{0.8} &
  \textbf{22.0} &
  \textbf{13.3} &
  \textbf{10.2} &
  \textbf{12.6} &
  \textbf{8.7} &
  \textbf{7.8} &
  \textbf{25.2} &
  \textbf{15.6} &
  \textbf{14.1} \\ \hline
\end{tabular}
\vspace{0.1cm}
\caption{Top-1 classification accuracy results on EPIC-Kitchens dataset under clean data and single-source adversarial perturbations on each modality. Higher is better. Due to space constraints, we defer Top-5 accuracy to Supplementary Materials. 
}
\label{table:adversarial-results-epic-kitchens}
\end{table*}

\begin{table*}[t]
\scriptsize
\centering
\begin{tabular}{|l|c|c|c|c|c|c|c|c|c|c|c|c|}
\hline
\textbf{Fusion} &
  \multicolumn{3}{c|}{\textbf{Clean}} &
  \multicolumn{3}{c|}{\textbf{\begin{tabular}[c]{@{}c@{}}Visual (RGB)\\ Perturbation\end{tabular}}} &
  \multicolumn{3}{c|}{\textbf{\begin{tabular}[c]{@{}c@{}}Depth (Velo)\\ Perturbation\end{tabular}}} &
  \multicolumn{3}{c|}{\textbf{\begin{tabular}[c]{@{}c@{}}Depth (Stereo)\\ Perturbation\end{tabular}}} \\ \hline
\textbf{} &
  \textbf{Car} &
  \textbf{Pedest.} &
  \textbf{Cyclist} &
  \textbf{Car} &
  \textbf{Pedest.} &
  \textbf{Cyclist} &
  \textbf{Car} &
  \textbf{Pedest.} &
  \textbf{Cyclist} &
  \textbf{Car} &
  \textbf{Pedest.} &
  \textbf{Cyclist} \\ \hline
\textbf{Oracle (Upper Bound)} &
 - &
 - &
 - &
 90.4 &
 80.1 &
 86.4 &
 93.2 &
 79.3 &
 85.3 &
 92.8 &
 80.5 &
 87.4
   \\ \hhline{|=|=|=|=|=|=|=|=|=|=|=|=|=|}
\textbf{Concat Fusion} &
  93.5 &
  \textbf{81.5} &
  \textbf{87.7} &
  14.3 &
  10.7 &
  12.3 &
  1.58 &
  11.1 &
  8.82 &
  3.57 &
  4.64 &
  7.23 \\ \hline
\textbf{Mean Fusion} &
  \textbf{93.6} &
  77.7 &
  86.7 &
  12.6 &
  15.2 &
  10.5 &
  3.16 &
  12.9 &
  7.88 &
  3.08 &
  8.03 &
  7.77 \\ \hhline{|=|=|=|=|=|=|=|=|=|=|=|=|=|}
\textbf{LEL+Robust \cite{kim2019single}} &
  71.4 &
  64.2 &
  80.0 &
  3.95 &
  15.4 &
  13.9 &
  6.83 &
  20.6 &
  24.8 &
  9.39 &
  24.2 &
  24.7 \\ \hline
\textbf{Gating+Robust \cite{kim2018robust,kim2018robusta}} &
  89.4 &
  74.7 &
  84.6 &
  57.2 &
  54.2 &
  56.0 &
  46.5 &
  45.7 &
  45.6 &
  41.6 &
  47.4 &
  48.8 \\ \hline
\textbf{Ours} &
  90.6 &
  79.9 &
  85.4 &
  \textbf{85.1} &
  \textbf{73.9} &
  \textbf{82.3} &
  \textbf{87.8} &
  \textbf{71.1} &
  \textbf{85.8} &
  \textbf{89.8} &
  \textbf{76.8} &
  \textbf{84.7} \\ \hhline{|=|=|=|=|=|=|=|=|=|=|=|=|=|}
\textbf{$\Delta$-Clean} &
  -3.0 &
  -1.6 &
  -2.3 &
  \textbf{70.8} &
  \textbf{58.7} &
  \textbf{70.0} &
  \textbf{74.6} &
  \textbf{58.2} &
  \textbf{77.0} &
  \textbf{86.2} &
  \textbf{68.8} &
  \textbf{76.9} \\ \hline
\textbf{$\Delta$-Robust} &
  \textbf{1.2} &
  \textbf{5.2} &
  \textbf{0.8} &
  \textbf{27.9} &
  \textbf{19.7} &
  \textbf{26.3} &
  \textbf{41.3} &
  \textbf{25.4} &
  \textbf{40.2} &
  \textbf{48.2} &
  \textbf{29.4} &
  \textbf{35.9} \\ \hline
\end{tabular}
\caption{Evaluation of Average Precision for 2D object detection on the KITTI dataset under clean data and single-source adversarial perturbations on each modality. Higher is better. Due to space constraints, only performance at medium difficulty is shown. See Supplementary Materials for full table with easy/medium/hard difficulties.}
\label{table:adversarial-results-kitti}
\end{table*}

\begin{table*}[t]
\begin{center}
\centering
\scriptsize
\begin{tabular}{|l|c|c|c|c|c|c|c|c|}
\hline
\textbf{Fusion} & \multicolumn{2}{c|}{\textbf{Clean}} & \multicolumn{2}{c|}{\textbf{\begin{tabular}[c]{@{}c@{}}Audio\\ Perturbation\end{tabular}}}  & \multicolumn{2}{c|}{\textbf{\begin{tabular}[c]{@{}c@{}}Video\\ Perturbation\end{tabular}}}  & \multicolumn{2}{c|}{\textbf{\begin{tabular}[c]{@{}c@{}}Text\\ Perturbation\end{tabular}}}  \\
\hline
\textbf{} &
  \textbf{2-class} &
  \textbf{7-class} &
    \textbf{2-class} &
  \textbf{7-class} &
    \textbf{2-class} &
  \textbf{7-class} &
    \textbf{2-class} &
  \textbf{7-class} \\
  \hline
\textbf{Oracle (Upper Bound)} & - & - & 78.64 & 49.10 & 73.36 & 47.84 & 69.82 & 40.28 \\ \hhline{|=|=|=|=|=|=|=|=|=|}

\textbf{Concat Fusion} & 79.82 & 49.69 & 56.92 & 21.38 & 51.23 & 19.75 & 39.50 & 9.97\\ \hline
\textbf{Mean Fusion} & 78.09 & 46.14& 52.63 & 20.75 & 49.37 & 17.02 & 35.50 & 8.88 \\ \hhline{|=|=|=|=|=|=|=|=|=|}
\textbf{LEL+Robust \cite{kim2019single}} & 79.09 & 49.92& 69.21 & 39.51 & 63.15 & 35.17 & 58.14 & 21.23 \\ \hline
\textbf{Gating+Robust \cite{kim2018robust, kim2018robusta}} &78.82 & 46.37&69.31 & 38.26&64.23 & 31.88&59.39 & 25.14 \\\hline
\textbf{Ours}    &\textbf{82.03} & \textbf{50.89}&\textbf{73.18} & \textbf{42.06}&\textbf{69.94} & \textbf{38.20}& \textbf{66.13} & \textbf{30.20}\\\hhline{|=|=|=|=|=|=|=|=|=|}
\textbf{$\Delta$-Clean} &\textbf{2.21}&\textbf{1.20}&\textbf{16.26}&\textbf{20.68}&\textbf{18.71}&\textbf{18.45}&\textbf{26.53}&\textbf{20.23}\\\hline
\textbf{$\Delta$-Robust} &\textbf{1.94}&\textbf{0.97}&\textbf{3.87}&\textbf{2.55}&\textbf{5.71}&\textbf{3.03}&\textbf{6.74}&\textbf{5.06}\\\hline
\end{tabular}
\end{center}
\caption{Binary and seven-class classification results (\%) on MOSI. Higher is better. Random guess is 50\% for binary and 14.3\% for seven class classification.}
\label{table:adversarial-results-mosi}
\end{table*}

\section{Experiments}\label{sec:benchmarks}

We evaluate the single-source adversarial robustness of multimodal models on three benchmark tasks: action recognition on EPIC-Kitchens, 2D object detection on KITTI, and sentiment analysis on MOSI. Existing benchmark tasks for studying single source corruptions in multimodal models have primarily focused on the object detection task with two modalities \cite{kim2019single, kim2018robust, kim2018robusta}. The benchmarks that we consider involve three input modalities and span a larger variety of tasks and data sources, ensuring generality of the conclusions drawn. A summary can be found in Table \ref{table:experiments}.

\subsection{Multimodal Benchmark Tasks}

\paragraph{Action recognition on EPIC-Kitchens.}
EPIC-Kitchens is the largest egocentric video dataset consisting of 39,596 video clips \cite{Damen2018EPICKITCHENS}. The objective is to predict the action taking place in the video, which is composed of one verb and one noun out of 126 and 331 classes respectively. 
%
Three modalities are available from the original dataset: visual information (RGB frames), motion information (optical flow), and audio information. 

\paragraph{Object Detection on KITTI.} 
KITTI is an autonomous driving dataset \cite{KITTI_OD} that contains stereo camera and LIDAR information for 2D object detection, where the objective is to draw bounding boxes around objects of interest from predefined classes, \ie, car, pedestrian, cyclist, etc. 
%
Existing works use different combinations and processed versions of the available data modalities for object detection. For the proposed benchmark, we consider the following three modalities based on common use in the literature: (1) RGB frames, which are used by the majority of detection methods, (2) LIDAR points projected to a sparse depth map and (3) a depth map estimated from the stereo views \cite{wang2019pseudo}. 
%

\paragraph{Sentiment Analysis on CMU-MOSI.}
 Multimodal Opinion-level Sentiment Intensity Corpus (CMU-MOSI) \cite{zadeh2016mosi} is a multimodal dataset for sentiment analysis consisting of 93 video clips of movie reviews, each of which are divided into an average of 23.2 segments. 
Each segment 
is labeled with a continuous sentiment intensity between $[-3,3]$. The objective is to predict the sentiment on a binary scale (\ie, negative v. positive) or 7-class scale (\ie, rounding to the nearest integer). MOSI contains three modalities: text, video and audio. 

\subsection{Implementation Details}

\paragraph{Model Architecture and Training.} For each task, we consider mid- to late- multimodal models that use the architectures summarized in column 4 of Table \ref{table:experiments}. We first train baseline multimodal models for each task on clean data to obtain $f_{\text{naive}}$. We then augment these models with the odd-one-out network and robust feature fusion layer as described in Section \ref{sec:method} to obtain $f_\text{robust}$, and perform robust training according to Algorithm \ref{alg:training}. Additional details are deferred to Supplementary Materials.

\paragraph{Adversarial Attacks.} The adversarial perturbations for each task are summarized in column 5 of Table \ref{table:experiments}. We attack individual modalities using projected gradient descent (PGD) \cite{madry2017towards}, except text, for which we use word replacement \cite{text_attack2}. Note that these perturbations are white-box adaptive attack, \ie, attacks are generated with full knowledge of $f_{\text{robust}}$. In the Supplementary Materials, we also describe and show results for other types of attacks, such as transfer attacks, targeted attacks, and feature-level attacks \cite{xu2020towards}.

\paragraph{Evaluation Metric.} The metrics used for each task are summarized in column 6 of Table \ref{table:experiments}. For the action recognition, we consider classification accuracy of verbs, nouns, and actions. For object detection, we consider the average precision of car, pedestrian, and cyclist detection at intersection-over-union (IoU) thresholds shown in the table, and at three difficulty levels following the KITTI evaluation server \cite{KITTI_OD}. For sentiment analysis, we consider binary and 7-class prediction accuracy. For each metric, we consider clean performance as well as performance under single-source attacks.

\subsection{Baselines}
In addition to our approach, we evaluate two types of methods: standard multimodal models trained with clean data (standard training), and state-of-the-art robust multimodal models \cite{kim2019single, kim2018robust, kim2018robusta} with robust training.

\paragraph{Concatenation Fusion with Standard Training (``Concat Fusion'').}  We use multimodal models with the same feature extractors and concatenate features before the final layers, which is a standard method for fusing features.

\paragraph{Mean Fusion with Standard Training (``Mean Fusion'').} For each modality, we train a unimodal model with the same feature extractor and final layers as the multimodal model on clean data. Then we fuse the unimodal model outputs by taking their mean, \ie, $z_{\text{output}} = \sum_{i \in [k]} z_i$. For action recognition and sentiment analysis, we perform mean fusion on the logits layer. For object detection, we perform the fusion prior to the YOLO layer. Mean fusion is a common fusion practice used in late fusion models, and in the context of defenses against perturbations, it is equivalent to a soft voting strategy between the different modalities.

\paragraph{Latent Ensembling Layer with Robust Training (``LEL+Robust'' \cite{kim2019single} ).} This approach involves (1) training on clean data and data with each single-source corruption in an alternating fashion, and (2) ensembling the multimodal features using concatenation fusion followed by a linear network. We adapt their strategy to our model by training our multimodal models with their fusion layer on data augmented with single-source perturbations.
    
\paragraph{Information-Gated Fusion with Robust Training (``Gating+Robust''\cite{kim2018robust, kim2018robusta}).}
This approach applies a multiplicative gating function to features from different modalities before ensembling them. 
The adaptive gating function is trained on clean data and data with single-source corruptions.
We adapt their robustness strategy to our models by training our multimodal models with their gated feature fusion layer on data augmented with single-source adversarial perturbations.

\paragraph{Upper Bound (``Oracle (Upper Bound)'').} To obtain an empirical upper bound for robust performance under attacks against each modality, we train and evaluate 2-modal models that exclude the perturbed modality. We refer to this model as the oracle because it assumes perfect knowledge of which modality is attacked (\ie, a perfect odd-one-out network), which is not available in practice. 

\section{Results}\label{sec:results}
\paragraph{How robust are standard multimodal models to single-source perturbations?} 
A key motivation for building multimodal models that fuse features from several modalities, beyond improving model performance, is to improve the robustness of the model to perturbations at any given modality. To this end, we first ask how well multimodal models trained on clean data that use concatenation fusion (a standard mid-fusion approach) or mean fusion (a standard late fusion approach) fare against a worst-case perturbation on any single modality. Since these models utilize features from three input modalities ($k=3$), we hypothesized that ensembling the perturbed features from one modality with the clean features from two other modalities could boost the robust performance of the model, at least compared to a unimodal model that receives the same attack.

Our empirical results suggest that standard multimodal models trained on clean data that use concatenation fusion (``Concat Fusion'') or mean fusion (``Mean Fusion'') are surprisingly vulnerable against single-source adversarial perturbations. Across the benchmark tasks, we observe drastic drops in performance of both types of models when any one of the modalities receives a worst-case perturbation (see Rows 2-3 of Table \ref{table:adversarial-results-epic-kitchens}, \ref{table:adversarial-results-kitti}, \ref{table:adversarial-results-mosi}). For the more challenging tasks with larger output spaces, such as action recognition on EPIC-Kitchens and object detection on KITTI, the multimodal models are not significantly better than a unimodal model under the same attack, and their performance is often close to zero.
In the Supplementary Materials, we show that similar results hold when we use other types of adversarial perturbations, such as transfer attacks, targeted attacks, and feature-level attacks \cite{xu2020towards}. An alarming conclusion drawn from the unimodal transfer attacks is that an attacker can successfully perturb a single modality of the multimodal model even without knowledge of how the different modalities are fused or what the inputs from unperturbed modalities are. 
%
%
%
Overall, these results show that standard multimodal fusion practices are not sufficiently robust against worst-case perturbations on a single modality and demonstrate the need for robust strategies.

%


\paragraph{How effective is the proposed approach at defending against single-source adversaries?}
%
%
%
Our proposed approach focuses on designing and training a robust feature fusion, using odd-one-out learning to leverage the correspondence between unperturbed modalities to detect and defend against any perturbed modality. We achieve significant gains in single-source adversarial robustness across all benchmarks and tasks under \textit{white-box adaptive attacks}, \ie, perturbations were generated with full knowledge of $f_{\text{robust}}$. The main results are shown in Tables \ref{table:adversarial-results-epic-kitchens}, \ref{table:adversarial-results-kitti}, \ref{table:adversarial-results-mosi}. Across the board, our method significantly improves the robustness of the standard models (see \textbf{``$\Delta$-Clean''}). Our approach also \emph{significantly outperforms} the state-of-the-art robust fusion methods, including the ``Gating+Fusion'' method which combines an adaptive gating function with robust training (see \textbf{``$\Delta$-Robust''}). In the Supplementary Materials, we show that our fusion strategy is also more robust against other single-source perturbations such as transfer attacks, targeted attacks, and feature-level attacks \cite{xu2020towards}. Comparing our method to the ``Oracle (Upper Bound)'' row, which always fuses the modalities that are unperturbed, we can see that our method is within -20\% of this empirical upper bound. We conclude that our approach outperforms the state-of-the-art in robust multimodal fusion and is close to empirical upper bound that can be achieved without learning robust features using end-to-end adversarial training.

\begin{table}[t]
\scriptsize
\centering
\begin{tabular}{|l|c|c|c|c|}
\hline
\multicolumn{5}{|c|}{\textbf{Action Recognition on EPIC-Kitchens}}                                                                                                         \\ \hline
\begin{tabular}[l]{@{}l@{}}Odd-one-out\\ network \end{tabular} &
  Clean &
  \begin{tabular}[c]{@{}c@{}}Visual\\ Perturb \end{tabular} &
  \begin{tabular}[c]{@{}c@{}}Motion\\ Perturb\end{tabular} &
  \begin{tabular}[c]{@{}c@{}}Audio\\ Perturb\end{tabular} \\ \hline
Unaligned features                                                          & \textbf{66.8} & \textbf{73.4} & \textbf{88.6} & \textbf{84.7} \\ \hline
Aligned Features         & 55.9          & 54.7 & 41.3          & 52.8          \\ \hline
\multicolumn{5}{|c|}{\textbf{Object Detection on KITTI}}                                                                                                                 \\ \hline
\begin{tabular}[l]{@{}l@{}}Odd-one-out\\ network \end{tabular} &
  Clean &
  \begin{tabular}[c]{@{}c@{}}RGB\\ Perturb\end{tabular} &
  \begin{tabular}[c]{@{}c@{}}Velo\\ Perturb\end{tabular} &
  \begin{tabular}[c]{@{}c@{}}Stereo\\ Perturb\end{tabular} \\ \hline
Unaligned features                                                          & \textbf{96.2} & \textbf{93.5} & \textbf{98.2} & \textbf{98.0} \\ \hline
Aligned Features & 91.9          & 86.8          & 94.4          & 90.4          \\ \hline
\multicolumn{5}{|c|}{\textbf{Sentiment Analysis on MOSI}}                                                                                                                 \\ \hline
\begin{tabular}[l]{@{}l@{}}Odd-one-out\\ network \end{tabular} &
  Clean &
  \begin{tabular}[c]{@{}c@{}}Audio\\ Perturb\end{tabular} &
  \begin{tabular}[c]{@{}c@{}}Video\\ Perturb\end{tabular} &
  \begin{tabular}[c]{@{}c@{}}Text\\ Perturb\end{tabular} \\ \hline
Unaligned features                                                          & \textbf{94.8} & \textbf{95.3} & \textbf{91.2} & \textbf{86.4}  \\ \hline
Aligned Features & 80.3                       & 90.4  & 87.3   & 78.5          \\ \hline
\end{tabular}
\caption{Detection rate (\%) of odd-one-out networks that use unaligned vs. aligned representations of features from each modality. Higher is better. Random guess is 25\%.}
\label{table:detection_rate}
\end{table}

\paragraph{Detection accuracy of odd-one-out network.}
We hypothesize that our approach is effective because odd-one-out learning enables the model to compare features from the different modalities, recognize consistent information between the unperturbed modalities, and exclude any perturbed modality that is inconsistent with the others. 
To determine if this is the case, we ask how well the odd-one-out network performs in detecting adversarial perturbations from each modality. 
The results in Table \ref{table:detection_rate} (see \textbf{``Unaligned features''}) suggest that the odd-one-out network is highly effective at filtering out features from the perturbed modalities, and performs better when there is more redundant information between the two unperturbed modalities. 

In relative comparison between the three tasks, we observe that the odd-one-out network is more successful at filtering out perturbed modalities on the KITTI and MOSI benchmarks than on the EPIC-Kitchens benchmark. This is consistent with our observation that the three modalities in the KITTI benchmark are highly redundant, which enables the odd-one-out network to detect the perturbed modality more easily. For the MOSI benchmark, the text modality contains the most information for the task, and audio and vision provide partly redundant information with text, which is also reflected in the results. In contrast, the three modalities in the EPIC-Kitchens benchmark (\eg, vision, motion, audio) contain a relatively higher degree of complementary (\ie, non-redundant) information, which increases the difficulty of odd-one-out learning. For example, if there is less shared information between the motion and audio inputs for a particular sample, then it may be harder to detect that the visual input is inconsistent.

Since the performance of the odd-one-out network in Table \ref{table:detection_rate} translates to the robust performance of the full model in Tables \ref{table:adversarial-results-epic-kitchens}, \ref{table:adversarial-results-kitti}, \ref{table:adversarial-results-mosi}, we asked if we could improve the detection accuracy using features that are already aligned, rather than the unaligned representations from individual feature extractors.
For example, for action recognition and sentiment analysis, one can compute differences between logits output by unimodal models; similarly, for object detection, one can compute differences between bounding box coordinates and object confidences output by unimodal models. We found that such strategies (Table \ref{table:detection_rate}, \textbf{``Aligned features''}) were generally less effective than using the unaligned features. 
This suggests that the context information available in the unaligned features from the feature extractors may be helpful for detecting the perturbed modality.
We defer the full robust performance of our models based on different odd-one-out networks (including a random baseline) to the Supplementary Material.

\paragraph{What are the advantages over learning robust features for each modality (\ie, end-to-end adversarial training)?} \emph{Clean performance:} Robust features that are trained on perturbed data are known to perform \textit{significantly} worse on unperturbed data \cite{madry2017towards}. In contrast, our approach is built on top of feature extractors pretrained on clean data and does not notably degrade clean performance. Across all three benchmarks, our performance on clean (unperturbed) data is comparable with fusion models with standard training (see the ``Clean'' column of the ``$\Delta$-Clean'' rows in Tables \ref{table:adversarial-results-epic-kitchens}, \ref{table:adversarial-results-kitti}, \ref{table:adversarial-results-mosi}). 
%
\emph{Resource utilization:} End-to-end adversarial training on perturbed data is resource-intensive \cite{adversarial_training_free} and may not be feasible for large multimodal models. In contrast, our Algorithm \ref{alg:training} for robust fusion only requires training the parameters in the odd-one-out network and the fusion network and not the parameters in the feature extractors.
%
This can drastically reduce the number of parameters that need to be trained on perturbed data. Table \ref{table:parameter_count} shows the number of parameters in the feature extractors vs. the fusion network for our different models. Note that on KITTI, our approach achieves single-source robustness by robustly training only $\sim$3.3\% of the model parameters.

\begin{table}[]
\centering
\scriptsize
\begin{tabular}{|l|c|c|}
\hline
\textbf{}              & \multicolumn{2}{c|}{\textbf{\# Parameters (Approx in Millions)}} \\ \hline
\textbf{\begin{tabular}[c]{@{}l@{}}Task\\ \end{tabular}} &
  \textbf{\begin{tabular}[c]{@{}c@{}}Feature Extractors\\ (Not Trained)\end{tabular}} &
  \textbf{\begin{tabular}[c]{@{}c@{}}Fusion Network\\ (Trained)\end{tabular}} \\ \hline
\textbf{EPIC-Kitchens} & 30.8                            & 57.9                           \\ \hline
\textbf{KITTI}         & 201.1                           & 6.8                            \\ \hline
\textbf{CMU-MOSI}      & 253.4                               & 12.3                             \\ \hline
\end{tabular}
\caption{Number of parameters (in millions) in the feature extractors and fusion networks of our multimodal models.}
\label{table:parameter_count}
\end{table}

\section{Conclusion}
This paper presents, to our knowledge, the first empirical study of multimodal robustness under single-source worst-case (adversarial) perturbations. 
We show across multiple benchmarks that standard multimodal fusion models are surprisingly vulnerable to single-source adversaries and provide an effective solution based on a robust feature fusion that leverages multimodal consistency through odd-one-out learning. The methods and experiments are focused on digital attacks on a single modality in the case where $k \geq 3$. Important directions for future work include physically-realisable attacks, as well as attacks on multiple modalities at once, which would require scaling up our robust fusion approach and training strategy. We believe that this first work on single-source adversarial attack and defense can serve as a basis for these future directions.

{\small
\bibliographystyle{ieee_fullname}
\bibliography{egbib}
}

\clearpage


\section*{Supplementary material}

# Supplementary Materials: Defending Multimodal Fusion Models against Single-Source Adversaries

Karren Yang[1]    Wan-Yi Lin[2]    Manash Barman[3]    Filipe Condessa[2]    Zico Kolter[2,3]
[1]Massachusetts Institute of Technology    [2]Bosch Center for AI    [3]Carnegie Mellon University

## A. Experimental Details

### A.1. Model Architecture and Training

**Action Recognition on Epic-Kitchens.** We adapt the state-of-the-art temporal binding networks by Kazakos et al. [3]. This model extracts features from each of the three modalities using an Inception network with Batch Normalization [12]. The features from different modalities are ensembled using feed-forward networks with ReLU activation prior to temporal aggregation and prediction of a verb and noun class. We obtain a pretrained model on clean data following the data preprocessing, augmentation, and training steps of Kazakos et al. [3]. We then augment the model with the robust fusion strategy described in Section 3 of the main text and perform training according to Algorithm 1 using stochastic gradient descent with momentum 0.9 and weight decay $5 \times 10^{-4}$ for ∼120 epochs with a learning rate of $1 \times 10^{-3}$.

**Object Detection on KITTI.** We adapt a single-shot detector based on YOLOv2 [9] and YOLOv3 [10] to the multimodal setting. The model extracts features from each of the three modalities using a Darknet19 network [9]. The features are fused using a $1 \times 1$ convolutional layer prior to predicting bounding boxes and confidences using a YOLO detector layer [10]. We obtain the Velodyne depth map by projecting the Velodyne points to the 2D image plane using the calibration and projection matrices provided in the dataset [2], followed by dilation with a $5 \times 5$ diamond kernel and bilateral smoothing. We obtain the stereo depth image using the pretrained PSMNet from [14]. All visual inputs are resized to $1280 \times 380$ for training and evaluation. During training, we augment the visual inputs by adding a random horizontal flip and a random shift that is at most one-fifth of the original width and height of the image, and we additionally augment the RGB image by randomly shifting the hue, exposure, and saturation. We obtain a pretrained multimodal model on unperturbed data in two steps: (i) we pretrain single-shot detectors for each modality using the same architecture, except for the fusion layer using stochastic gradient descent with momentum 0.9 and weight decay $5 \times 10^{-4}$ for ∼500 epochs with a learning rate of $1 \times 10^{-3}$, (ii) we train the fusion layer over the pretrained Darknet19 networks with the same optimization for ∼120 epochs. We then augment the model with the robust fusion strategy described in Section 3 of the main text and perform training according to Algorithm 1 using stochastic gradient descent (SGD) with momentum 0.9, weight decay $5 \times 10^{-4}$, and gradient clip of 20 for ∼20 epochs with a learning rate of $1 \times 10^{-3}$.

**Sentiment Analysis on CMU-MOSI.** For video input, we use VGGFace2 [1] followed by one layer of multi-head attention [13] and two layers of bi-directional LSTMs; for audio, we apply the same architecture on the Mel-frequency cepstral coefficients (MFCCs) of the audio signal instead of VGGFace2 outputs; for text, we apply the Transformer model [13] on a 300 dimensional pretrained GloVe embedding trained on Wikipedia from [8]. We first train unimodal models using each of these three feature extractor followed by a fully-connected layer. We then augment the model with the robust fusion strategy described in Section 3 of the main text and perform training according to Algorithm 1 using stochastic gradient descent with momentum 0.9, weight decay $5 \times 10^{-4}$, a learning rate of $1 \times 10^{-5}$, and trained for 40 epochs.

### A.2. Adversarial Perturbations

The adversarial perturbations considered in our experiments are white box adaptive attack, *i.e.*, attacks are generated with full knowledge of $f_{\text{robust}}$.

**Action Recognition on EPIC-Kitchens.** We consider single-source $\ell_\infty$ attacks on each of three modalities: vision, motion, and audio. To generate the perturbations, we approximate the solution of Equation 1 of the main text using projected gradient descent (PGD) [7], taking $\mathcal{L}$ to be the sum of the cross-entropy losses from the verb and noun classes. For vision and motion, we approximate the solution with 10-step PGD with $\epsilon = 8/256$. For the audio spectrogram, we approximate the solution with 10-step PGD with $\epsilon = 0.8$.

**Object Detection on KITTI.** We consider single-source $\ell_\infty$ perturbations on each of three visual modalities. To generate the perturbations, we approximate the solution

| Odd-one-out network | Clean | | | Visual Perturbation | | | Motion Perturbation | | | Audio Perturbation | | |
|---|---|---|---|---|---|---|---|---|---|---|---|---|
| | Verb | Noun | Action | Verb | Noun | Action | Verb | Noun | Action | Verb | Noun | Action |
| Random | 59.6 | **43.1** | 30.1 | 0.0 | 0.0 | 0.0 | 0.0 | 0.0 | 0.0 | 0.0 | 0.0 | 0.0 |
| Unaligned features | 61.5 | 42.5 | **31.4** | **48.0** | **24.2** | **16.8** | **48.5** | **35.6** | **21.1** | **46.5** | **33.3** | **22.1** |
| Aligned features (Logits) | **61.5** | 38.0 | 31.2 | 38.7 | 19.5 | 13.1 | 37.0 | 29.1 | 16.7 | 31.1 | 23.9 | 13.8 |

Table 1. We show the top-1 classification accuracy of our fusion model with different odd-one-out networks on action recognition on the EPIC-Kitchens dataset under clean data and single-source adversarial perturbations on each modality. Higher is better.

| Odd-one-out network | Clean | | | Visual (RGB) Perturbation | | | Depth (Velo) Perturbation | | | Stereo Disparity Perturbation | | |
|---|---|---|---|---|---|---|---|---|---|---|---|---|
| | Car | Pedest. | Cyclist | Car | Pedest. | Cyclist | Car | Pedest. | Cyclist | Car | Pedest. | Cyclist |
| Random | 90.1 | 79.0 | 84.3 | 17.9 | 7.6 | 7.6 | 1.8 | 19.8 | 28.1 | 1.1 | 18.4 | 30.2 |
| Unaligned features | **90.6** | **79.9** | **85.4** | **85.1** | **73.9** | **82.3** | **87.8** | **71.1** | **85.8** | **89.8** | **76.8** | **84.7** |
| Aligned features (Bounding Boxes) | 90.2 | 76.1 | 83.3 | 77.6 | 66.5 | 73.8 | 83.1 | 64.8 | 68.3 | 80.0 | 65.1 | 44.9 |

Table 2. We show the average precision (at medium difficulty) of our fusion model with different odd-one-out networks on object detection on the KITTI dataset under clean data and single-source adversarial perturbations on each modality. Higher is better.

| Odd-one-out network | Clean | | Audio Perturbation | | Video Perturbation | | Text Perturbation | |
|---|---|---|---|---|---|---|---|---|
| | 2-class | 7-class | 2-class | 7-class | 2-class | 7-class | 2-class | 7-class |
| Random | 75.41 | 46.73 | 74.48 | 33.52.10 | 61.65 | 31.46 | 62.98 | 25.01 |
| Unaligned features | **82.03** | **50.89** | **73.18** | **42.06** | **69.94** | **38.20** | **66.13** | **30.20** |
| Aligned features | 80.13 | 48.26 | 71.95 | 39.74 | 66.47 | 35.43 | 60.74 | 26.57 |

Table 3. We show the binary and 7-class accuracy of our fusion model with different odd-one-out networks on sentiment analysis on the CMU-MOSI dataset under clean data and single-source adversarial perturbations on each modality. Higher is better.

of Equation 1 in the main text using 10-step PGD with $\epsilon = 16/256$, taking $\mathcal{L}$ to be the YOLO loss [10]. The loss consists of the cross-entropy losses of object and class confidences as well as the mean-squared localization errors summed over bounding box proposals.

**Sentiment Analysis on CMU-MOSI.** We consider single-source attacks on each of three modalities: vision, audio, and text. The adversarial perturbations approximate the solution of Equation 1 in the main text, taking $\mathcal{L}$ to be the binary or 7-class cross entropy. For vision, we generate $\ell_\infty$ perturbations with 10-step PGD with $\epsilon = 8/256$. For audio, we generate the single-source $\ell_\infty$ perturbations with 10-step PGD with $\epsilon = 0.8$. For text, we adapted the word-replacement attack with priority based on word saliency [11] with an attack budget of one word per sentence.

### A.3. Evaluation Metrics

For each evaluation metric, we consider clean performance (*i.e.*, performance on unperturbed data) as well as robust performance on data with single-source adversarial perturbations. For robust performance, we report the evaluation metric once for each attacked modality.

**Action Recognition on EPIC-Kitchens.** Models are evaluated based on their top-1 and top-5 classification accuracy on the verb, noun, and action classes.

**Object Detection on KITTI.** We consider Average Precision (AP) on three object classes: cars, pedestrians, and cyclists. For the car class, true positives are bounding boxes with Intersection over Union (IoU) > 0.7 with real boxes. For the pedestrian and cyclist classes, true positives are bounding boxes with IoU > 0.5 with real boxes.

**Sentiment Analysis on CMU-MOSI.** Models are evaluated based on their binary classification accuracy (*i.e.*, predicting whether the sentiment is positive or negative) as well as their classification accuracy of 7 sentiment classes.

## B. Robust performance with different odd-one-out networks

In Table 5 of the main text, we show that the odd-one-out network based on unaligned representations of different features is more effective at detecting the perturbed modality than a random baseline as well as an odd-one-out network based on aligned representations of the features. Supplementary Tables 1, 2 and 3 show task performance of the models using different odd-one-out networks. Overall, we observe that task performance reflects the detection rates; when the model's odd-one-out network is more effective at detecting the perturbed modality, it is more successful at only allowing information from the unperturbed modalities to pass through the feature fusion step, resulting in better task performance under single-source perturbations. Specifically, we show that the model with the odd-one-out network based on unaligned representations of different features is more robust than the baselines that use a random odd-one-out network or an odd-one-out network based on aligned representations of the features.

| Fusion | Clean | | | Visual Perturbation | | | Motion Perturbation | | | Audio Perturbation | | |
|---|---|---|---|---|---|---|---|---|---|---|---|---|
| | Verb | Noun | Action | Verb | Noun | Action | Verb | Noun | Action | Verb | Noun | Action |
| Oracle (Upper Bound) | - | - | - | 55.8 | 31.4 | 21.9 | 50.0 | 37.2 | 23.8 | 53.9 | 39.2 | 25.6 |
| Concat Fusion | 59.0 | 42.1 | 30.2 | 1.3 | 0.8 | 0.3 | 0.5 | 1.7 | 0.1 | 1.5 | 2.1 | 0.4 |
| Mean Fusion | 56.8 | 40.4 | 27.6 | 2.3 | 0.8 | 0.4 | 0.4 | 1.3 | 0.1 | 1.9 | 2.5 | 0.5 |
| LEL+Robust [6] | 61.2 | 43.1 | 30.5 | 54.5 | 30.4 | 21.1 | 51.4 | 36.6 | 22.7 | 51.5 | 37.3 | 23.9 |
| Gating+Robust [5, 4] | 60.9 | 43.0 | 30.6 | 56.9 | 32.5 | 22.7 | 52.0 | 39.9 | 25.2 | 55.6 | 39.4 | 26.3 |
| Ours | **61.5** | 42.5 | **31.4** | **58.0** | **33.9** | **24.5** | **53.2** | 39.2 | **25.6** | **56.6** | **39.5** | **27.1** |
| Δ-Clean | 2.5 | 0.3 | 1.2 | 55.7 | 33.1 | 24.1 | 52.7 | 37.5 | 25.5 | 54.7 | 37.0 | 26.6 |
| Δ-Robust | 0.3 | -0.6 | 0.8 | 1.1 | 1.4 | 1.8 | 1.2 | -0.7 | 0.4 | 1.0 | 0.1 | 0.8 |

Table 4. **Unimodal transfer attack.** Top-1 classification accuracy results on EPIC-Kitchens dataset under clean data and single-source adversarial perturbations on each modality, where the adversarial perturbations are generated and transferred from unimodal models. Higher is better.

| Fusion | Clean | | | Visual Perturbation | | | Motion Perturbation | | | Audio Perturbation | | |
|---|---|---|---|---|---|---|---|---|---|---|---|---|
| | Verb | Noun | Action | Verb | Noun | Action | Verb | Noun | Action | Verb | Noun | Action |
| Oracle (Upper Bound) | - | - | - | 55.8 | 31.4 | 21.9 | 50.0 | 37.2 | 23.8 | 53.9 | 39.2 | 25.6 |
| Concat Fusion | 59.0 | 42.1 | 30.2 | 0.0 | 0.0 | 0.0 | 0.3 | 0.1 | 0.0 | 0.0 | 0.0 | 0.0 |
| Mean Fusion | 56.8 | 40.4 | 27.6 | 0.0 | 0.0 | 0.0 | 0.2 | 0.0 | 0.0 | 0.0 | 0.0 | 0.0 |
| LEL+Robust [6] | 55.8 | 36.7 | 23.9 | 9.4 | 7.2 | 3.1 | 11.3 | 17.8 | 5.1 | 6.2 | 12.3 | 3.1 |
| Gating+Robust [5, 4] | 57.1 | 39.1 | 26.6 | 18.8 | 12.6 | 6.7 | 32.3 | 25.7 | 14.2 | 16.3 | 13.0 | 7.6 |
| Ours | **60.2** | **42.5** | **31.0** | **53.6** | **29.7** | **20.8** | **48.7** | **35.6** | **22.7** | **50.7** | **36.4** | **23.5** |
| Δ-Clean | 1.2 | 0.4 | 0.8 | 53.5 | 29.7 | 20.8 | 48.4 | 35.5 | 22.7 | 50.7 | 36.4 | 23.5 |
| Δ-Robust | 3.1 | 3.4 | 4.4 | 34.8 | 17.1 | 14.1 | 16.4 | 9.9 | 8.5 | 34.4 | 23.4 | 15.9 |

Table 5. **Feature-level attack.** Top-1 classification accuracy results on EPIC-Kitchens dataset under clean data and single-source adversarial perturbations on each modality, where the adversarial perturbations are generated at the feature-level, rather than the input level, as proposed in [15]. Higher is better.

| Fusion | Clean | | | Visual Perturbation | | | Motion Perturbation | | | Audio Perturbation | | |
|---|---|---|---|---|---|---|---|---|---|---|---|---|
| | Verb | Noun | Action | Verb | Noun | Action | Verb | Noun | Action | Verb | Noun | Action |
| Oracle (Upper Bound) | - | - | - | 55.8 | 31.4 | 21.9 | 50.0 | 37.2 | 23.8 | 53.9 | 39.2 | 25.6 |
| Concat Fusion | 59.0 | 42.1 | 30.2 | 15.9 | 5.5 | 2.0 | 18.8 | 8.4 | 2.6 | 15.9 | 8.1 | 2.6 |
| Mean Fusion | 56.8 | 40.4 | 27.6 | 23.9 | 12.4 | 5.3 | 25.3 | 14.3 | 5.3 | 22.6 | 13.4 | 4.0 |
| LEL+Robust [6] | **62.3** | 42.3 | 31.0 | 31.6 | 23.1 | 6.7 | 38.5 | 32.2 | 14.8 | 31.5 | 28.7 | 9.6 |
| Gating+Robust [5, 4] | 61.9 | **44.7** | **31.9** | 35.7 | 25.9 | 9.3 | 45.9 | 36.2 | 19.5 | 38.2 | 28.8 | 11.8 |
| Ours | 60.8 | 43.5 | 31.3 | **45.1** | **32.1** | **17.5** | **51.5** | **38.0** | **24.5** | **56.6** | **40.0** | **27.7** |
| Δ-Clean | 1.8 | 1.4 | 1.1 | 21.2 | 19.7 | 12.2 | 26.2 | 23.7 | 19.2 | 34.0 | 26.6 | 23.7 |
| Δ-Robust | -1.5 | -1.2 | -0.6 | 9.4 | 6.2 | 8.2 | 5.6 | 1.8 | 5.0 | 18.4 | 11.2 | 15.9 |

Table 6. **Untargeted attack.** Top-1 classification accuracy results on EPIC-Kitchens dataset under clean data and single-source adversarial perturbations on each modality, where the adversarial perturbations are generated to target the verb and noun class of a randomly sampled clip from the dataset, rather than the real verb and noun class. Higher is better.

## C. Other Types of Attacks

In the main text of the paper, we showed across three benchmark tasks that standard multimodal models are not robust to single-source adversaries (*e.g.*, end-to-end, untargeted white box attacks on the entire multimodal model) and subsequently proposed a defense strategy based on odd-one-out learning for robust feature fusion. Here, we show on the EPIC-Kitchens dataset that our results also hold for other types of attacks: transfer attacks, targeted attacks, and feature-level attacks [15].

**Transfer attacks.** We consider unimodal transfer attacks on the multimodal models by (1) training a unimodal classifier on top of pretrained feature extractors for each modality, (2) generating a white-box untargeted attack from this unimodal model, (3) applying the perturbed input to the multimodal model along with the two unperturbed inputs from the other modalities. Note that the perturbed input is generated without knowledge of the fusion model and without knowledge of the unperturbed inputs from other modalities. From the performance results of different models in Supplementary Table 4, we draw two conclusions. First, standard multimodal models based on concatenation fusion ("Concat Fusion") or mean fusion ("Mean Fusion") are also not robust to these transfer attacks, which are weaker than end-to-end white box attacks; in fact, the performance is close to zero, which is no better than a unimodal model defending against the same attack. In other words, an adversary can successfully attack a single modality of a multimodal model without knowledge of how features are fused between different modalities and without knowledge of the unperturbed inputs from other modalities. Second, our approach out-

performs the standard models and existing state-of-the-art robust methods on these types of attacks (see "Δ-Clean", "Δ-Robust"). We note, however, that all robust methods perform reasonably well against transfer attacks, as they are not adaptive attacks.

**Feature-level attacks.** Feature level attacks, as proposed in [15], are a good alternative way to assess the robustness of the fusion model, since here the attacker does not use the vulnerability of the feature extractors to generate the attack. To perform a feature-level attack, we perform a white-box attack on the feature representation for a particular modality after it has been extracted by a feature extractor $g_i$ and before it has been fused using $h$ (or $\tilde{h}$). We consider feature-level attacks with a maximum perturbation strength of $\epsilon = 3$ for each modality. The performance results of different models are shown in Supplementary Table 5. We note that standard multimodal models are also vulnerable to feature-level attacks; in fact, the performance is close to zero, which is no better than a unimodal model defending against the same attack. Additionally, our approach significantly outperforms the standard models and existing state-of-the-art robust methods on these types of attacks (see "Δ-Clean", "Δ-Robust").

**Targeted attacks.** To perform a targeted attack, we first randomly select a verb and noun class from the dataset and denote this target label as $\hat{y}$; we denote the predicted verb and noun for the input **x** as $y$. Subsequently, we generate a white-box adversarial attack against the model by optimizing Equation (1) of the main text, taking

$$\mathcal{L}(f(x_i + \delta, \mathbf{x}_{-i}), \hat{y}) = -\log \mathbb{P}_f(y = \hat{y}; x_i + \delta, \mathbf{x}_{-i}).$$

The performance results of different models are shown in Supplementary Table 6. Importantly, our approach also outperforms the standard models and existing state-of-the-art robust methods on these types of attacks (see "Δ-Clean", "Δ-Robust"). Note that the standard models appear to perform better than zero here; however, this is mainly because there is a chance of sampling targeted verb and noun classes that are the same as the real labels.

## D. Clean performance of robust fusion strategy on KITTI

Features that are trained on perturbed data are known to perform *significantly* worse on unperturbed data [7]. In contrast, our approach is built on top of feature extractors pre-trained on clean data and does not notably degrade clean performance. In the main text, we show across all three benchmarks that our performance on clean (unperturbed) data is comparable with fusion models with standard training (see the "Clean" column of the "Δ-Clean" rows in Tables 2,3,4 of the main text). Clean performance seems slightly degraded on the KITTI dataset, but these differ-

| Fusion | Clean Performance | | |
|---|---|---|---|
| | **Car** | **Pedestrian** | **Cyclist** |
| **Non-Robust Model** | 92.8 ± 1.3 | 81.0 ± 4.7 | 84.9 ± 4.7 |
| **Ours** | 91.3 ± 2.7 | 80.8 ± 5.0 | 86.2 ± 4.5 |

Table 7. Clean performance of our robust model v. standard multimodal model, trained for the same number of epochs. The uncertainties shown are standard deviations computed over three splits of the validation data.

| | **Verb** | **Noun** | **Action** |
|---|---|---|---|
| **Vision** | 39.6 / 83.1 | 35.4 / 60.8 | 18.7 / 54.2 |
| **Motion** | 53.2 / 83.4 | 28.6 / 53.5 | 19.0 / 48.4 |
| **Audio** | 40.7 / 77.3) | 20.6 / 40.9 | 12.1 / 35.9 |
| **V+M** | 53.9 / 85.4 | 39.2 / 64.6 | 25.6 / 58.2 |
| **V+A** | 50.0 / 84.3 | 37.2 / 63.4 | 23.8 / 55.9 |
| **M+A** | 55.8 / 84.6 | 31.4 / 57.8 | 21.9 / 52.2 |
| **V+M+A** | 59.0 / 86.1 | 42.1 / 66.5 | 30.2 / 60.8 |

Table 8. **EPIC-Kitchens Multimodal Benchmark Task**. Top-1 / Top-5 classification accuracy of standard 1-modal, 2-modal, and 3-modal models on the Epic-Kitchens verb, noun, and action recognition task. Rows 2-4 show the performance of 1-modal models based on individual modalities; rows 5-7 show the performance of 2-modal models on every pair of modalities, and row 8 shows the performance of the 3-modal model on all of the modalities.

ences are not significant when standard deviations are taken into account as shown in Supplementary Table 7.

## E. Analysis of Multimodal Benchmark Tasks

In order to leverage consistency between modalities in a multimodal task to perform robust fusion, there must be shared or redundant information between them. Here, we analyze the extent to which the modalities in EPIC-Kitchens, KITTI, and CMU-MOSI are redundant, by considering the clean performance of 1-modal (unimodal), 2-modal, and 3-modal models on the task. The results in this section help shed light on the discussion of the odd-one-out network performance in the main text. Specifically, it is easier for the odd-one-out network to detect the perturbed modality for datasets such as KITTI where there is large overlap in information between data modalities; on the other hand, odd-one-out learning is more challenging on datasets such as EPIC-Kitchens where there is a fair amount of complementary information between data modalities.

**EPIC-Kitchens Action Recognition.** The results of our assessment of the EPIC-Kitchens multimodal benchmark are shown in Supplementary Table 8. We observe a trend of diminishing returns on clean performance as we add more modalities to the model, which indicates that there is redundant information between the modalities— notice that the drop of performance is relatively small from the 3-modal models (row 8) to any of the 2-modal models (rows 5-7), especially for the verb and noun classification tasks. How-

|  | Car | Pedestrian | Cyclist |
|---|---|---|---|
| **RGB** | 93.3 / 92.9 / 86.0 | 82.3 / 76.3 / 71.9 | 90.3 / 83.0 / 80.4 |
| **Velodyne** | 89.6 / 87.0 / 81.0 | 82.2 / 76.2 / 72.1 | 91.7 / 84.6 / 81.8 |
| **Stereo Depth** | 93.9 / 88.3 / 81.6 | 79.8 / 71.7 / 67.8 | 83.8 / 80.4 / 75.7 |
| **R+V** | 93.3 / 92.8 / 86.2 | 85.0 / 80.5 / 76.3 | 95.1 / 87.4 / 85.1 |
| **R+S** | 96.1 / 93.2 / 86.5 | 84.6 / 79.3 / 75.0 | 90.3 / 85.3 / 82.7 |
| **V+S** | 93.5 / 90.4 / 83.8 | 87.0 / 80.1 / 76.0 | 93.4 / 86.4 / 84.0 |
| **R+V+S** | 96.1 / 93.5 / 86.8 | 85.8 / 81.5 / 77.0 | 93.2 / 87.7 / 83.2 |

Table 9. **Kitti Multimodal Benchmark Task**. Average Precision of standard 1-modal, 2-modal, and 3-modal models on Easy / Medium / Hard object detection as defined by the official KITTI evaluation server [2]. Rows 2-4 show the performance of 1-modal models based on individual modalities; rows 5-7 show the performance of 2-modal models on every pair of modalities, and row 8 shows the performance of the 3-modal model on all of the modalities.

|  | Two-class | Seven class |
|---|---|---|
| **Video** | 69.09 | 39.32 |
| **Audio** | 59.32 | 33.01 |
| **Text** | 75.23 | 46.64 |
| **V+A** | 69.82 | 40.28 |
| **V+T** | 78.64 | 49.10 |
| **A+T** | 73.36 | 47.84 |
| **V+A+T** | 79.82 | 49.69 |

Table 10. **CMU-MOSI sentiment analysis Benchmark Task**. 2-class/7-class sentiment classification accuracy (%) of standard 1-modal, 2-modal, and 3-modal models. Rows 2-4 show the performance of 1-modal models based on individual modalities; rows 5-7 show the performance of 2-modal models on every pair of modalities, and row 8 shows the performance of the 3-modal model on all of the modalities.

ever, there is also a notable amount of complementary information between the modalities. For example, the visual modality contains more information for noun classification: the 1-modal vision model outperforms the 1-modal motion and audio models in noun accuracy, and the drop from the 3-modal model to the 2-modal model excluding vision yields the largest drop in noun accuracy. Likewise, the motion modality contains more information for verb classification: the 1-modal motion model outperforms the 1-modal vision and audio models in verb accuracy, and the drop from the 3-modal model to the 2-modal model excluding motion yields the largest drop in verb accuracy. Our findings are consistent with the observations in [3].

**KITTI 2D Object Detection.** The results of our assessment of the KITTI multimodal benchmark are shown in Supplementary Table 9. Here, we observe that each modality achieves high performance on the task individually. The improvement of clean performance when increasing the number of modalities from 1 (rows 2-4) to 3 (row 8) is rather minimal. This demonstrates that the information between these three modalities is highly redundant, and there is little complementary information between the modalities.

**CMU-MOSI Sentiment Analysis.** The results of our assessment of CMU-MOSI benchmark are shown in Supplementary Table 10. In this task, text is the modality that carries most information and adding video improves the performance more significantly than adding audio. Comparing row 8 which uses all three modalities and row 7 which uses video and text, as well as comparing row 4 which uses video and audio with row 1 which uses only video, one can see that there exist strong redundancy between audio and video.

## F. Early Fusion vs. Late Fusion Models

In our preliminary experiments, we found that early fusion models are notably less robust than late fusion methods, which motivated us to focus on late fusion approaches in this work. Supplementary Table 11 shows our results comparing early and late fusion approaches on action recognition on the UCF-101 dataset. Here the input consists of 2-5 frames (which are treated as different views) and only one of the frames is adversarially perturbed while the others are clean An early fusion method that concatenates the frames over their spatial dimensions before passing them through a convolutional neural network is significantly less robust than a late fusion network that processes the frames individually and averages the output.

We also evaluated the robustness of transformers that perform early fusion of video, audio, and text from the MOSI dataset, and found that they performed notably worse than late fusion models as shown in Supplementary Table 12.

One intuitive explanation for these results is the following: in early fusion models, there are multiple early interactions between attacked and clean modalities, so feature extraction becomes strongly influenced by the attacked input. In contrast, in late fusion models, feature extraction of the clean modalities is mostly shielded from the attack.

## G. Additional Tables

In the following, we provide additional results that were deferred from the main text due to space constraints. Sup-

|  |  | **Number of Input Frames** | | | |
|---|---|---|---|---|---|
|  | **Input Type** | **2** | **3** | **4** | **5** |
| **Early Fusion Model** | Clean frames | 69.7/89.3 | 70.2/89.1 | 68.3/87.2 | 70.9/87.4 |
|  | Single-frame attack | 16.3/38.1 | 20.8/45.1 | 20.0/45.0 | 22.5/48.8 |
| **Late Fusion Model** | Clean frames | 72.4/89.0 | 73.0/90.0 | 74.0/89.6 | 73.5/90.5 |
|  | Single-frame attack | 46.9/76.4 | 62.0/83.5 | 66.7/85.8 | 67.0/87.4 |

Table 11. **Early fusion vs. late fusion robustness on UCF-101 action recognition.** The early fusion model is significantly less robust than the late fusion model under a single-frame perturbation. Top 1 / Top 5 classification accuracy is shown (higher is better). See text for details.

|  | Audio | Video | Text |
|---|---|---|---|
| **Early Fusion (transformer) Model** | 15.9 | 18.2 | 27.5 |
| **Late Fusion Model** | 56.92 | 51.23 | 39.1 |

Table 12. **Early fusion vs. late fusion robustness on two-class MOSI sentiment analysis.** The early fusion model is significantly less robust than the late fusion model, especially under audio and video perturbation.

plementary Table 13 provides top-1 and top-5 accuracy of multimodal models on action recognition on EPIC-Kitchens and supplement the result of Table 2 in the main text. Supplementary Table 14 shows average precision performance of multimodal models on easy/medium/hard object detection on the KITTI dataset and supplement the results of Table 3 in the main text. Supplementary Table 15 shows end-to-end adversarial training results on MOSI – the proposed method outperforms adversarial training on clean input for over 20% with small sacrifice on robust accuracy.

| Fusion | Clean | | | Visual Perturbation | | | Motion Perturbation | | | Audio Perturbation | | |
|---|---|---|---|---|---|---|---|---|---|---|---|---|
| | Verb | Noun | Action | Verb | Noun | Action | Verb | Noun | Action | Verb | Noun | Action |
| Concat Fusion | 59.0 (86.1) | 42.1 (66.5) | 30.2 (60.8) | 0.1 (25.0) | 0.0 (7.3) | 0.0 (4.0) | 0.2 (24.6) | 0.0 (5.1) | 0.0 (2.7) | 0.1 (21.2) | 0.0 (6.8) | 0.0 (3.5) |
| Mean Fusion | 56.8 (85.9) | 40.4 (66.1) | 27.6 (59.7) | 0.3 (58.1) | 0.8 (15.8) | 0.0 (11.8) | 0.3 (64.2) | 0.3 (14.3) | 0.0 (12.0) | 0.4 (48.0) | 0.3 (16.9) | 0.0 (12.0) |
| LEL+Robust [6] | 61.2 (86.7) | **43.1 (68.9)** | 30.5 (62.3) | 22.3 (76.3) | 11.6 (37.2) | 6.6 (34.9) | 25.4 (78.6) | 24.6 (52.3) | 12.0 (46.5) | 20.4 (77.6) | 17.7 (47.5) | 8.0 (43.2) |
| Gating+Robust [5, 4] | 60.9 (87.6) | 43.0 (68.7) | 30.6 (62.8) | 26.0 (77.9) | 10.9 (42.0) | 6.2 (39.0) | 35.9 (83.6) | 26.9 (58.3) | 14.3 (52.5) | 21.3 (79.9) | 16.2 (51.4) | 7.0 (47.2) |
| Ours | **61.5 (88.2)** | 42.5 (68.4) | **31.4 (63.0)** | **48.0 (83.2)** | **24.2 (53.2)** | **16.8 (48.9)** | **48.5 (85.2)** | **35.6 (63.8)** | **22.1 (57.4)** | **46.5 (85.2)** | **33.3 (62.3)** | **22.1 (57.1)** |

Table 13. Top-1 (Top-5) classification accuracy results on EPIC-Kitchens dataset under clean data and single-source adversarial perturbations on each modality. Higher is better. See Table 2 in main paper and accompanying details in text for details.

| Fusion | Clean | | | Visual (RGB) Perturbation | | |
|---|---|---|---|---|---|---|
| | Car | Ped | Cyc | Car | Ped | Cyc |
| Concat Fusion | 96.1 / 93.5 / 86.8 | 85.7 / 81.5 / 77.0 | 93.2 / 87.7 / 83.2 | 15.6 / 14.3 / 14.8 | 13.6 / 10.7 / 10.3 | 13.8 / 12.3 / 12.9 |
| Mean Fusion | 96.5 / 93.6 / 86.6 | 84.3 / 77.7 / 73.4 | 91.9 / 86.7 / 81.8 | 13.2 / 12.6 / 13.1 | 18.1 / 15.2 / 14.2 | 11.9 / 10.5 / 10.2 |
| LEL+Robust [6] | 80.5 / 71.4 / 67.3 | 69.0 / 64.2 / 61.3 | 75.7 / 80.0 / 75.3 | 3.71 / 3.95 / 4.62 | 16.9 / 15.4 / 14.3 | 16.4 / 13.9 / 13.2 |
| Gating+Robust [5, 4] | 90.6 / 89.4 / 82.8 | 81.5 / 74.7 / 72.6 | 92.9 / 84.6 / 81.8 | 67.3 / 57.2 / 53.1 | 62.0 / 54.2 / 50.7 | 68.6 / 56.0 / 53.1 |
| Ours | 95.6 / 90.6 / 83.9 | 84.5 / 79.9 / 75.7 | 90.4 / 85.4 / 80.6 | **89.6 / 85.1 / 78.9** | **80.5 / 73.9 / 69.8** | **87.9 / 82.3 / 77.6** |
| Fusion | Depth (Velo) Perturbation | | | Stereo Disparity Perturbation | | |
| | Car | Ped | Cyc | Car | Ped | Cyc |
| Concat Fusion | 3.43 / 1.58 / 1.59 | 11.3 / 11.1 / 11.4 | 8.72 / 8.82 / 8.22 | 7.37 / 3.57 / 3.44 | 8.08 / 4.64 / 4.36 | 9.13 / 7.23 / 7.72 |
| Mean Fusion | 6.77 / 3.16 / 2.90 | 13.7 / 12.9 / 12.8 | 10.1 / 7.88 / 8.07 | 6.88 / 3.08 / 2.73 | 9.17 / 8.03 / 8.81 | 12.2 / 7.77 / 7.28 |
| LEL+Robust [6] | 7.30 / 6.83 / 6.73 | 24.4 / 20.6 / 18.9 | 28.8 / 24.8 / 24.7 | 10.1 / 9.39 / 9.50 | 26.0 / 24.2 / 21.8 | 25.7 / 24.7 / 23.8 |
| Gating+Robust [5, 4] | 51.7 / 46.5 / 43.2 | 53.5 / 45.7 / 42.1 | 58.8 / 45.8 / 53.8 | 43.9 / 41.6 / 38.8 | 53.9 / 47.4 / 44.1 | 60.0 / 48.8 / 46.8 |
| Ours | **92.8 / 87.8 / 79.4** | **78.3 / 71.1 / 67.1** | **88.9 / 85.8 / 81.1** | **92.8 / 89.8 / 83.1** | **83.6 / 76.8 / 72.4** | **88.1 / 84.7 / 79.9** |

Table 14. Evaluation of Average Precision for 2D object detection (easy/medium/hard difficulty) on the KITTI dataset under clean data and single-source adversarial perturbations on each modality. Higher is better. See Table 3 in main paper and accompanying details in text for details.

| Fusion | Clean | | Audio Perturbation | | Video Perturbation | | Text Perturbation | |
|---|---|---|---|---|---|---|---|---|
| | 2-class | 7-class | 2-class | 7-class | 2-class | 7-class | 2-class | 7-class |
| Adversarial training | 61.23 | 36.02 | 74.38 | 42.13 | 70.34 | 39.28 | 69.94 | 32.45 |
| Ours | **82.03** | **50.89** | 73.18 | 42.06 | 69.94 | 38.20 | 66.13 | 30.20 |

Table 15. Binary and seven-class classification results (%) of end-to-end adversarial training and our method on MOSI. Higher is better.


\end{document}